\documentclass[10pt,twocolumn]{article}

\usepackage[preprint]{antintlpaper}
\usepackage[utf8]{inputenc}
\usepackage{inconsolata}
\usepackage{adjustbox}
\usepackage{algorithm}
\usepackage{algpseudocode}
\usepackage{makecell}
\usepackage{xurl}
\usepackage{fontawesome5}
\usepackage{hyperref}

\fancypagestyle{plain}{%
  \fancyhf{}
  \fancyhead[L]{\sffamily\fontsize{7.8}{9}\selectfont\bfseries ANT INTERNATIONAL RESEARCH}
  \fancyhead[R]{\sffamily\fontsize{7.8}{9}\selectfont Preference Adaptation via Verbal RL}
  \fancyfoot[C]{\sffamily\fontsize{8}{9}\selectfont\thepage}
  \renewcommand{\headrulewidth}{0.5pt}
  \renewcommand{\headrule}{\hbox to\headwidth{\color{AntRule}\leaders\hrule height \headrulewidth\hfill}}
}

\AntTitle{Learning Preference Adaptation for Large Language Model Personalization via Verbal Reinforcement Learning}
\AntRunningTitle{Learning Preference Adaptation for Large Language Model Personalization via Verbal Reinforcement Learning}
\AntAuthors{%
  Yuting Liu\textsuperscript{1,2} \and
  Wei Wu\textsuperscript{2,*} \and
  Jianzhe Zhao\textsuperscript{1} \and
  Guibing Guo\textsuperscript{1,*}}
\AntAffiliations{%
  \textsuperscript{1}Software College, Northeastern University, China
  \quad
  \textsuperscript{2}Ant International}
\AntContact{%
  \textsuperscript{*}Corresponding authors: Guibing Guo and Wei Wu.\\
  \texttt{liuyuting@stumail.neu.edu.cn},
  \texttt{wuwei19850318@gmail.com},
  \texttt{\{guogb,zhaojz\}@swc.neu.edu.cn}}
\AntDate{}
\AntKeywords{large language model personalization, task-specific preference adaptation, verbal reinforcement learning, meta-learning}
\AntAbstract{Natural language user preferences provide an interpretable interface for LLM personalization. However, universal preference summaries often contain information irrelevant to a particular downstream task. Directly supplying the full preference summary therefore wastes context capacity and introduces cross-task distraction, while manually designing task-specific preference views is difficult to scale. In this work, we study \emph{task-specific preference adaptation}: given a universal user preference summary and a downstream task, derive a task-conditioned representation that preserves sufficient decision-relevant evidence while removing redundant context. To this end, we propose \textsc{AlignXada}, a training-free meta-learning framework that induces reusable textual refinement policies for adapting universal preference summaries to task-specific ones. The refinement policy is iteratively optimized by a meta learner through verbal reinforcement learning. Across 13 tasks and three downstream models (39 task--model cells), \textsc{AlignXada} achieves an average gain of 3.82 points, improving 33 cells while retaining only 22.8\% of the original profile tokens and outperforming RAG in 36 cells. An extended faithfulness analysis further shows that the refined profiles remain largely grounded in the source preferences while preserving task-relevant personalization signals, suggesting that profile-side adaptation serves as a practical complement to universal memory construction for lifelong personalized agents. Code is available at \url{https://github.com/AntResearchNLP/AlignX-Family/tree/main/AlignXada}.
}

\hypersetup{pdfauthor={Yuting Liu, Wei Wu, Jianzhe Zhao, Guibing Guo}}

\begin{document}

\makeanttitle

\section{Introduction}
\label{sec:intro}

Personalization, the technique of aligning AI systems with human preferences, has played an important role in the training of large language models since their early development \citep{instructGPT}. Recently, with the proliferation of personal AI agents \citep{OpenClaw,HermesAgent} and the expanding use of LLMs in applications such as search \citep{baek2024knowledge}, recommendation \citep{geng2022recommendation,lyu2024llmrec, peng2026memrerank}, and dialogue systems~\citep{otsuka2024user}, personalization techniques --- particularly as a key component in building LLM memories~\citep{longmem,amem,mem0,yu2025memagent,memorybank} --- have become increasingly critical and continue to drive advances at the research frontier.

\begin{figure*}[tbp]
    \centering
    \includegraphics[width=.95\linewidth]{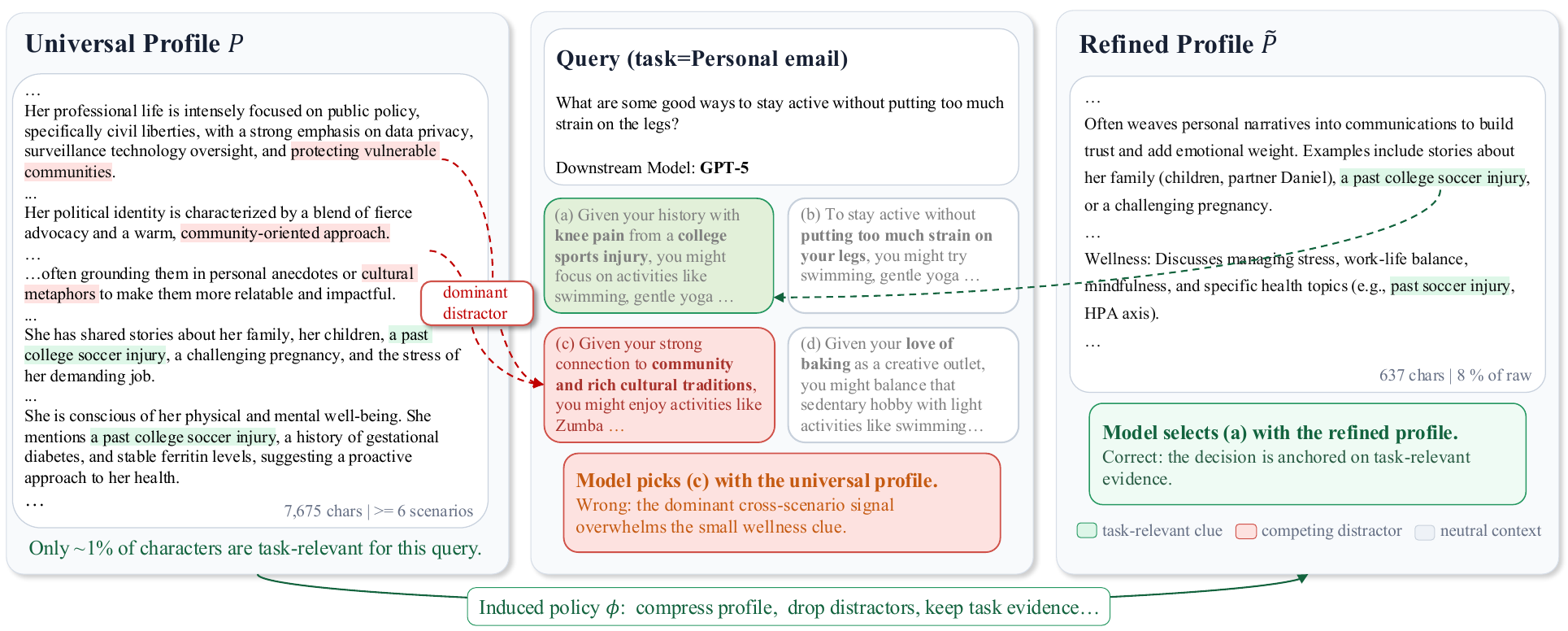}
    \caption{An example for task-specific preference adaptation. Given a query, only a small portion of the universal profile $P$ is relevant to the task, while the remaining information may act as distractors and lead to suboptimal personalization. \textsc{AlignXada} refines $P$ by retaining task-relevant evidence and removing distracting context, enabling the downstream model to ground its response in the appropriate user information.}
    \label{fig:example}
\end{figure*}

A central problem in personalization is user preference representation, which serves as the interface through which AI systems perceive user preferences and adapt their behaviors accordingly. Early studies modeled user preferences through user embeddings or by encoding user information directly into model parameters~\citep{mf,bao2023tallrec,Liu2024CoRACI,zhang2025stylevector}. While effective, such black-box representations suffer from limited interpretability and are difficult to update in real time. With the emergence of LLMs, in-context learning has enabled a simple yet effective alternative by explicitly incorporating user behaviors into the model context as demonstrations~\citep{purple,salemi2024optimization}. However, such approaches are fundamentally constrained by the context capacity of LLMs, making it difficult to comprehensively capture user preferences while remaining highly sensitive to noise in the selected examples. More recently, the advancement of LLM reasoning capabilities has inspired efforts to infer universal user preferences from heterogeneous data sources and summarize them in natural language~\citep{chen2025popi,nam2025plus,alignxploreplus,yang2026sensorpersonal}. 
This paradigm offers significant improvements in interpretability and scalability. More importantly, it lays the foundation for lifelong personal agents, where preference representations can continuously evolve as user--agent interactions accumulate over time.

In this work, we study LLM personalization from the perspective of universal user preference interfaces. Rather than pursuing improved methods for constructing universal preference summaries, we assume a universal user profile is already available and investigate a complementary yet practically important problem: \textit{how to effectively adapt the universal profile to specific tasks}. Our motivation stems from the observation that universal preference summaries, while comprehensive, inevitably contain redundant or task-irrelevant information that may act as noise in downstream applications. 
As illustrated in Figure~\ref{fig:example}, the user query is strongly associated with her past college injury, whereas information such as ``community-oriented approach'' is largely irrelevant. Such redundant information misleads the downstream model (i.e., GPT-5), resulting in a suboptimal response. In contrast, the result can be substantially improved by using an adapted profile that is shorter and more focused on task-relevant information.

Toward task-specific preference adaptation, we propose a framework for transforming universal preference summaries into task-aware representations. Ideally, the adapted representation should satisfy two desiderata: (1) \textit{Sufficiency}, preserving all preference information necessary for downstream personalization; and (2) \textit{Compactness}, removing redundant or task-irrelevant information to reduce noise and improve context efficiency. To this end, we propose \textsc{AlignXada}, a meta-learning framework for preference adaptation. Rather than directly training a model to refine user profiles, \textsc{AlignXada} employs a meta learner to learn structured refinement policies in natural language, which are subsequently used by a frozen refiner to adapt the universal user preference. This disentanglement between policy generation and preference refinement makes the adaptation process transparent and controllable, enabling human-in-the-loop diagnosis and refinement. The meta learner is optimized via verbal reinforcement learning, where refinement policies are iteratively improved using natural language feedback derived from task-specific demonstrations of user preferences. By avoiding parameter updates during policy learning, \textsc{AlignXada} naturally supports both open-source and proprietary models.

We evaluate \textsc{AlignXada} on a composite benchmark spanning nine conversational tasks and four ranking, rating, and generation tasks with three downstream models. Across 39 task--model cells, \textsc{AlignXada} improves 33 cells by an average of 3.82 points while reducing the profile token ratio to 22.8\%. It strictly outperforms RAG in 36 cells, showing that task-oriented preference reorganization provides benefits beyond query-level retrieval. On PersonaMem-v2, a faithfulness audit further shows that 97.5\% of refined-profile claims are supported by the source profiles, while 83.3\% of available gold preference evidence is retained, suggesting that \textsc{AlignXada} mainly performs controlled task-specific compression and reorganization.

We summarize our contributions as follows:
\begin{itemize}
\item We formalize the \emph{task-specific preference adaptation} problem, which aims to adapt universal user preferences to downstream tasks by removing redundant or task-irrelevant information, thereby paving the way for lifelong personalized agents equipped with memory.
\item We introduce \textsc{AlignXada}, a meta-learning framework that induces natural language refinement policies from a small set of task-specific demonstrations. The policy is iteratively optimized via verbal reinforcement learning, making \textsc{AlignXada} compatible with both open-source and proprietary LLMs.
\item We conduct extensive evaluations across thirteen tasks and three downstream models. The results show that \textsc{AlignXada} consistently achieves a favorable trade-off between task performance and context-token usage, owing to the faithfulness and compactness of the refined preferences.
\end{itemize}

\begin{figure*}
    \centering
    \includegraphics[width=0.95\linewidth]{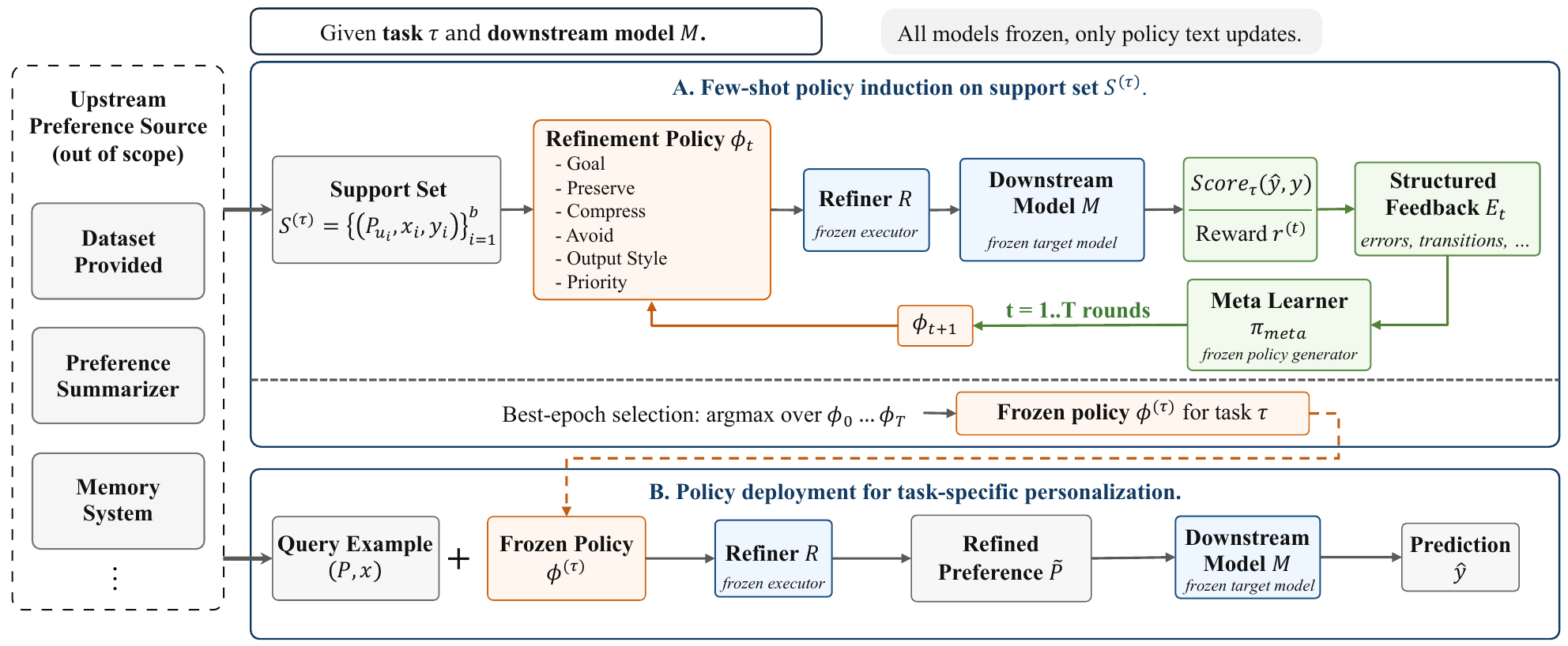}
    \caption{Overview of \textsc{AlignXada}. For each task, \textsc{AlignXada} induces a task-specific textual refinement policy from a small support set in several rounds and then freezes the selected policy for held-out deployment. All models remain frozen throughout, and only the policy text is updated during induction.}
    \label{fig:framework}
\end{figure*}
\section{Related Work}
\label{sec:related}

\subsection{LLM Personalization}
\label{sec:related-personalization}

As large language models (LLMs) evolve from generalized chatbots into personal AI agents, there has been growing interest in building personalized AI systems in which the behavior of a general-purpose LLM is aligned with individual preferences. Existing approaches can be broadly categorized into four groups. Retrieval-based methods identify relevant user records or profile elements from external memory or databases~\citep{personadb,purple,recap,cfrag}, and leverage the retrieved content for downstream personalization. Parametric approaches encode persona information into trainable model parameters, such as LoRA modules, adapters, soft prompts, model-merging weights, and rerankers~\citep{oppu,peftu,pepler,pplug,personalizedsoups,hydra}, thereby adapting a general-purpose LLM toward user-specific behavior. In addition, motivated by the strong in-context learning capability of LLMs, prompt-based approaches append preference signals to the input context and steer response generation through personalized prompts~\citep{dong2023steer,yang2024rewardsincontext,cheng2024indialogues,li20251000000usersuserscaling}. More recently, with the emergence of lifelong personal agents and LLM memory systems, several studies have advocated learning universal user profiles as long-term and continuously evolving representations of user preferences~\citep{li2025extendedinductivereasoningpersonalized,alignxploreplus,yang2026sensorpersonal}. \textsc{AlignXada} is motivated by this trend toward lifelong personalization. However, rather than constructing universal profiles themselves, \textsc{AlignXada} assumes such profiles are already available and studies how to adapt them to specific downstream tasks in real-world applications. In this sense, \textsc{AlignXada} complements existing efforts by bridging the gap between universal profile construction and task-oriented deployment.

\subsection{Textual Optimization}
\label{sec:related-textopt}

With advances in LLM reasoning capabilities, recent work has used strong LLMs to optimize natural-language artifacts without gradient-based training. OPRO~\citep{opro} iteratively proposes task-level instructions based on reward trajectories, EvoPrompt~\citep{evoprompt} applies evolutionary operators to candidate prompts, TextGrad~\citep{textgrad} propagates verbal ``gradients'' through prompt computation graphs, and Reflexion~\citep{reflexion} generates reflective critiques for per-instance trajectories. In these methods, the optimized artifact is typically either a \emph{task-level} instruction shared across users or a \emph{per-instance} self-correction text, making it user-independent or instance-local. \textsc{AlignXada} extends this paradigm by formulating task-specific profile adaptation as the optimization of a reusable \emph{user-conditional} rewrite policy that transforms each user's universal preference into a task-specific representation, rather than introducing a new general-purpose textual optimizer.

\section{Methodology}
\label{sec:method}
Figure~\ref{fig:framework} presents an overview of \textsc{AlignXada}. In a nutshell, \textsc{AlignXada} consists of two stages: few-shot policy induction and policy deployment for task-specific personalization. During policy induction, the framework leverages a small support set of task-specific demonstrations, each consisting of a universal preference summary, a user query, and the corresponding user response, and iteratively refines the rewriting policy generated by a meta learner using natural language feedback. Once policy optimization converges, \textsc{AlignXada} enters the deployment stage, where the selected policy is consumed by a refiner to produce a refined profile for task adaptation. The refined profile is then provided to downstream models for personalized inference. Throughout the entire process, all models—including the meta learner, the refiner, and the downstream model—remain frozen. In the following, Section~\ref{sec:method-formalization} formalizes the learning problem, and Section~\ref{sec:method-method} presents the policy induction procedure based on verbal reinforcement learning.

\subsection{Problem Formalization}
\label{sec:method-formalization}

Let $u$ denote a user and $P_u$ denote the corresponding universal preference summary, which may be obtained from an external LLM or memory system and is treated as prior knowledge in this work. Given a task $\tau$, the objective is to derive a task-adapted profile $\tilde{P}_u$ from $P_u$ such that the performance of a downstream model $M$ for user $u$ can be substantially improved on task $\tau$.

A common approach to deriving $\tilde{P}_u$ from $P_u$ is to learn a generative model $R$ as a refiner and define $\tilde{P}_u = R(P_u, \mathcal{M}_{\tau})$, where $\mathcal{M}_{\tau}$ specifies the task $\tau$. In this work, we instantiate $\mathcal{M}_{\tau}$ as a support set $S^{(\tau)}$ defined as
\begin{equation}
S^{(\tau)}=\{(P_{u_i}, x_i,y_i)\}_{i=1}^{b},
\end{equation}
where $P_{u_i}$ is the universal profile of user $u_i$, $x_i$ denotes an input prompt, $y_i$ denotes the corresponding response, and $b$ is the size of the support set. Although the representational capacity of $S^{(\tau)}$ may be limited by the selection of $(P_{u_i}, x_i,y_i)$ tuples and the budget $b$, such a formulation naturally aligns with real-world user-AI interactions: user prompts specify the task, while user responses provide signals of task-specific preferences and behavioral patterns, thereby alleviating the cold-start problem.

Instead of updating the parameters of $R$, we keep the model frozen and learn a textual refinement policy $\phi^{(\tau)}$ using a meta learner $\pi_{\mathrm{meta}}$. The task-adapted profile is then defined as $\tilde{P}_u = R(P_u,\phi^{(\tau)})$. The learning objective is to maximize
\begin{equation}
\label{obj}
\mathbb{E}_{(P_{u_i},x_i,y_i)\sim S^{(\tau)}} \left[\mathrm{Score}_{\tau} \big(M(x_i,\tilde{P}_{u_i} ), y_i \big)\right],
\end{equation}
where $\mathrm{Score}_{\tau}(\cdot,\cdot)$ denotes the evaluation function for task $\tau$. 

We propose a verbal reinforcement learning approach to optimize Eq.~(\ref{obj}), where the refinement policy $\phi^{(\tau)}$ is iteratively estimated from $\pi_{\mathrm{meta}}$ using feedback derived from $S^{(\tau)}$. Details are presented in  next section.

\subsection{Policy Induction}
\label{sec:method-method}

\begin{algorithm}[t]
\caption{Task-specific policy induction in \textsc{AlignXada}.}
\label{alg:MePrO}
\begin{algorithmic}[1]
\Require Task $\tau$; support set $S^{(\tau)}$; development set $D^{(\tau)}$; frozen meta learner $\pi_{\mathrm{meta}}$; frozen refiner $R$; frozen downstream model $M$; evaluation function $\mathrm{Score}_{\tau}$; initial policy $\phi_0$; number of rounds $T$
\Ensure Task-specific refinement policy $\phi^{(\tau)}$

\State $\mathcal{H} \gets \emptyset$
\Comment{history of evaluated policies}

\For{$t = 0, 1, \ldots, T$}
    \State $\mathcal{R}_t \gets \emptyset$
    \Comment{support-set rollout records}

    \ForAll{$({P}_{u_i}, x_i, y_i) \in S^{(\tau)}$}
        \State $\tilde{P}_{u_i,t} \gets R({P}_{u_i}, \phi_t)$
        \State $\hat{y}_{i,t} \gets M(x_i, \tilde{P}_{u_i,t})$
        \State $s_{i,t} \gets \mathrm{Score}_{\tau}(\hat{y}_{i,t}, y_i)$
        \State $\begin{aligned}
                \mathcal{R}_t &\gets \mathcal{R}_t \cup \\&\quad\{({P}_{u_i}, x_i, y_i, \tilde{P}_{u_i,t},\hat{y}_{i,t}, s_{i,t})\}
            \end{aligned}$
    \EndFor
    
    \State $J_{D^{(\tau)}}(\phi_t)\leftarrow\frac{1}{|D^{(\tau)}|}\sum_{i=1}^{|D^{(\tau)}|} s^{D}_{i,t}$

    \State $\mathcal{H} \gets \mathcal{H} \cup\{(\phi_t, J_{D^{(\tau)}}(\phi_t))\}$

    \If{$t < T$}
        \State $E_t \gets \mathrm{AGG}(\mathcal{R}_t)$
        \State $\phi_{t+1} \gets \pi_{\mathrm{meta}}(\phi_t, E_t)$
    \EndIf
\EndFor

\State $\phi^{(\tau)} \gets \arg\max_{(\phi_t, J_{D^{(\tau)}}(\phi_t)) \in \mathcal{H}} J_{D^{(\tau)}}(\phi_t)$

\State \Return $\phi^{(\tau)}$
\end{algorithmic}
\end{algorithm}

\paragraph{Overview.}
Algorithm~\ref{alg:MePrO} presents the task-specific policy induction procedure in \textsc{AlignXada}. Starting from a predefined task-agnostic initial policy $\phi_0$ (detailed in Appendix~\ref{app:phi0}), \textsc{AlignXada} iteratively induces a refinement policy for task $\tau$ through two core flows: \emph{rollout} and \emph{update}. At each round $t$, the rollout flow evaluates the current policy $\phi_t$ on the support set $S^{(\tau)}$ by applying the refiner $R$, querying the downstream model $M$, and computing task-specific scores. The update flow then summarizes the rollout records into structured feedback $E_t$, which is used by the meta learner $\pi_{\mathrm{meta}}$ to revise the policy and produce the next policy $\phi_{t+1}$. After all rounds, \textsc{AlignXada} returns the policy with the highest development-set performance.

\paragraph{Rollout.}
At round $t$, \textsc{AlignXada} evaluates the current refinement policy $\phi_t$ on each support example $(P_{u_i},x_i,y_i)\in S^{(\tau)}$. 
The refiner $R$ applies $\phi_t$ to rewrite the universal preference summary into a task-adapted profile:
\begin{equation}
  \tilde{P}_{u_i,t}=R(P_{u_i},\phi_t).
  \label{eq:rewrite}
\end{equation}
The detailed prompt template is provided in Appendix~\ref{app:rewrite-prompt}. The downstream model then conditions on the input $x_i$ and the refined profile $\tilde{P}_{u_i,t}$ to generate a prediction:
\begin{equation}
  \hat{y}_{i,t}=M(x_i,\tilde{P}_{u_i,t}).
\end{equation}
This prediction is evaluated against the reference response $y_i$ using the task-specific evaluation function:
\begin{equation}
  s_{i,t}=\mathrm{Score}_{\tau}(\hat{y}_{i,t},y_i).
  \label{eq:score}
\end{equation}
The rollout record at round $t$ is then defined as
\begin{equation}
  \mathcal{R}_t =
  \{(P_{u_i},x_i,y_i,\tilde{P}_{u_i,t},\hat{y}_{i,t},s_{i,t})\}_{i=1}^{|S^{(\tau)}|}.
\end{equation}
To reduce overfitting, \textsc{AlignXada} evaluates each policy on a development set $D^{(\tau)}$ disjoint from $S^{(\tau)}$. Applying Eqs.~\ref{eq:rewrite}--\ref{eq:score} to its examples yields development scores $s^{D}_{i,t}$, whose average defines the policy utility:
\begin{equation}
J_{D^{(\tau)}}(\phi_t)
=
\frac{1}{|D^{(\tau)}|}
\sum_{i=1}^{|D^{(\tau)}|}
s^{D}_{i,t}.
\end{equation}
The evaluated policy and its development-set utility are stored in the history:
\begin{equation}
  \mathcal{H}
  \leftarrow
  \mathcal{H}
  \cup
  \{(\phi_t,J_{D^{(\tau)}}(\phi_t))\}.
\end{equation}
Maintaining this history allows \textsc{AlignXada} to retain all evaluated policies, as verbal policy updates are not guaranteed to improve monotonically.

\paragraph{Policy update.}
After the rollout, \textsc{AlignXada} constructs structured feedback:
\begin{equation}
  E_t=\mathrm{AGG}(\mathcal{R}_t),
\end{equation}
where $\mathrm{AGG}(\cdot)$ converts the rollout records into a textual diagnostic summary. 
The feedback includes scalar signals, such as the average support-set score; instance-level signals, such as predictions and scores; and representative failure patterns derived from low-scoring examples. 
The complete feedback format is provided in Appendix~\ref{app:feedback-format}.

Given the current policy and structured feedback, the frozen meta learner produces a revised policy:
\begin{equation}
  \phi_{t+1}=\pi_{\mathrm{meta}}(\phi_t,E_t).
\end{equation}
The update prompt (provided in Appendix~\ref{app:meta-learner-prompt}) instructs $\pi_{\mathrm{meta}}$ to follow the predefined policy schema, make targeted revisions based on the feedback, and avoid example-specific rules that merely memorize support instances. 

The revised policy may adjust the refinement goal, the types of user evidence to preserve, the compression strategy, the error patterns to avoid, the output style, or the priority order among these instructions. 
An example policy is provided in Appendix~\ref{app:policy}.
In this way, the meta learner converts rollout diagnostics into a reusable task-level refinement policy rather than producing direct answers or per-example corrections.

After $T$ rounds, \textsc{AlignXada} selects and returns the best evaluated policy as the final task-specific refinement policy:
\begin{equation}
  \phi^{(\tau)}
  =
  {\arg\max}_{(\phi_t,J_{D^{(\tau)}}(\phi_t))\in\mathcal{H}} \:
  J_{D^{(\tau)}}(\phi_t).
\end{equation}

\section{Experiments}
\label{sec:experiments}

\begin{table*}[t]
  \centering
  \caption{Overall performance (\%) on the composite benchmark with different downstream models. Higher metric is better unless otherwise specified. Token ratio (TR) denotes the ratio of refined profile token length to those in the original profile. Task-level gains over Raw are significant under a two-sided exact sign test ($p=2.44\times10^{-4}$).}
  \label{tab:main-result}
  \adjustbox{width=\linewidth}{
  \begin{tabular}{ccccccccccccc}
    \toprule
    \textbf{Task} &
    \multicolumn{4}{c}{\textbf{Qwen3-8B}} &
    \multicolumn{4}{c}{\textbf{DeepSeek-V4-Flash}} &
    \multicolumn{4}{c}{\textbf{GPT-5-mini}} \\
    \cmidrule(lr){2-5}
    \cmidrule(lr){6-9}
    \cmidrule(lr){10-13}
    &
    \textbf{Raw} & \textbf{RAG} & \textbf{\textsc{AlignXada}} & {TR(\%)}$\downarrow$ &
    \textbf{Raw} & \textbf{RAG} & \textbf{\textsc{AlignXada}} & {TR(\%)}$\downarrow$ &
    \textbf{Raw} & \textbf{RAG} & \textbf{\textsc{AlignXada}} & {TR(\%)}$\downarrow$ \\
    \midrule
    \makecell[c]{\textbf{Chat}\\Acc.}           & 43.46 & 40.78$_{\mathrm{-2.68}}$ & \textbf{45.63}$_{\mathrm{+2.17}}$ & 18.9 & 55.34 & 41.75$_{\mathrm{-13.59}}$ & \textbf{64.08}$_{\mathrm{+8.74}}$ & 18.1 & 43.40 & 40.78$_{\mathrm{-2.62}}$ & \textbf{49.51}$_{\mathrm{+6.11}}$ & 19.6 \\
    \makecell[c]{\textbf{Creative}\\Acc.}       & 55.68 & 46.59$_{\mathrm{-9.09}}$ & \textbf{56.82}$_{\mathrm{+1.14}}$ & 21.7 & 59.09 & 54.55$_{\mathrm{-4.54}}$ & \textbf{73.86}$_{\mathrm{+14.77}}$ & 22.5 & 59.32 & 61.36$_{\mathrm{+2.04}}$ & \textbf{65.91}$_{\mathrm{+6.59}}$ & 28.9 \\
    \makecell[c]{\textbf{Knowledge}\\Acc.}      & 72.19 & 61.54$_{\mathrm{-10.65}}$ & \textbf{75.00}$_{\mathrm{+2.81}}$ & 14.3 & \textbf{82.90} & 71.01$_{\mathrm{-11.89}}$ & 82.84$_{\mathrm{-0.06}}$ & 33.5 & \textbf{75.21} & 71.01$_{\mathrm{-4.20}}$ & 73.63$_{\mathrm{-1.58}}$ & 18.7 \\
    \makecell[c]{\textbf{Personal}\\Acc.}       & \textbf{34.52} & 33.33$_{\mathrm{-1.19}}$ & 33.33$_{\mathrm{-1.19}}$ & 24.9 & 45.24 & 39.29$_{\mathrm{-5.95}}$ & \textbf{54.76}$_{\mathrm{+9.52}}$ & 20.7 & 47.62 & 41.67$_{\mathrm{-5.95}}$ & \textbf{50.00}$_{\mathrm{+2.38}}$ & 17.1 \\
    \makecell[c]{\textbf{Prof.Email}\\Acc.}     & \textbf{39.17} & 35.00$_{\mathrm{-4.17}}$ & 37.67$_{\mathrm{-1.50}}$ & 25.1 & 46.67 & 35.83$_{\mathrm{-10.84}}$ & \textbf{59.17}$_{\mathrm{+12.5}}$ & 25.8 & \textbf{41.50} & 38.33$_{\mathrm{-3.17}}$ & 40.83$_{\mathrm{-0.67}}$ & 26.4 \\
    \makecell[c]{\textbf{Prof.Writing}\\Acc.}   & 31.82 & 27.27$_{\mathrm{-4.55}}$ & \textbf{40.00}$_{\mathrm{+8.18}}$ & 14.6 & 50.00 & 40.91$_{\mathrm{-9.09}}$ & \textbf{57.27}$_{\mathrm{+7.27}}$ & 27.8 & 44.55 & 36.36$_{\mathrm{-8.19}}$ & \textbf{49.09}$_{\mathrm{+4.54}}$ & 16.8 \\
    \makecell[c]{\textbf{Social}\\Acc.(\%)}     & 36.36 & 34.09$_{\mathrm{-2.27}}$ & \textbf{40.91}$_{\mathrm{+4.55}}$ & 20.3 & 56.82 & 53.41$_{\mathrm{-3.41}}$ & \textbf{62.50}$_{\mathrm{+5.68}}$ & 37.9 & 48.41 & 48.86$_{\mathrm{+0.45}}$ & \textbf{51.14}$_{\mathrm{+2.73}}$ & 19.2 \\
    \makecell[c]{\textbf{Translation}\\Acc.}    & 38.64 & 34.23$_{\mathrm{-4.41}}$ & \textbf{42.24}$_{\mathrm{+3.60}}$ & 24.4 & 54.95 & 45.95$_{\mathrm{-9.00}}$ & \textbf{63.06}$_{\mathrm{+8.11}}$ & 18.6 & 54.95 & 50.45$_{\mathrm{-4.50}}$ & \textbf{59.46}$_{\mathrm{+4.51}}$ & 21.7 \\
    \makecell[c]{\textbf{Trouble}\\Acc.}        & 49.51 & 46.60$_{\mathrm{-2.91}}$ & \textbf{53.40}$_{\mathrm{+3.88}}$ & 22.3 & 60.19 & 55.34$_{\mathrm{-4.85}}$ & \textbf{62.14}$_{\mathrm{+1.94}}$ & 24.3 & \textbf{55.19} & \textbf{55.19}$_{\mathrm{+0.00}}$ & 54.37$_{\mathrm{-0.82}}$ & 18.5 \\
    \makecell[c]{\textbf{Ranking}\\Hit$@$1}     & 59.61 & 55.77$_{\mathrm{-3.84}}$ & \textbf{61.53}$_{\mathrm{+1.92}}$ & 20.8 & 69.23 & 78.85$_{\mathrm{+9.62}}$ & \textbf{82.69}$_{\mathrm{+13.46}}$& 31.9 & 78.69 & 76.92$_{\mathrm{-1.77}}$ & \textbf{80.77}$_{\mathrm{+2.08}}$ & 26.7 \\
    \makecell[c]{\textbf{Rating}\\Score}        & 75.84 & 75.48$_{\mathrm{-0.36}}$ & \textbf{77.40}$_{\mathrm{+1.56}}$ & 28.2 & 74.04 & 76.44$_{\mathrm{+2.40}}$ & \textbf{78.37}$_{\mathrm{+4.33}}$ & 27.2 & 77.88 & \textbf{80.29}$_{\mathrm{+2.41}}$ & 79.33$_{\mathrm{+1.45}}$ & 26.4 \\
    \makecell[c]{\textbf{Title}\\ROUGE-L}       & 13.46 & 12.99$_{\mathrm{-0.47}}$ & \textbf{15.38}$_{\mathrm{+1.92}}$ & 13.1 & 12.78 & 10.76$_{\mathrm{-2.02}}$ & \textbf{16.02}$_{\mathrm{+3.24}}$ & 20.4 & 13.04 & 13.17$_{\mathrm{+0.13}}$ & \textbf{13.80}$_{\mathrm{+0.76}}$ & 30.3 \\
    \makecell[c]{\textbf{Review}\\ROUGE-L}      & 13.94 & 13.90$_{\mathrm{-0.04}}$ & \textbf{14.37}$_{\mathrm{+0.43}}$ & 22.2 & 13.89 & 13.86$_{\mathrm{-0.03}}$ & \textbf{15.32}$_{\mathrm{+1.43}}$ & 22.8 & 11.61 & 12.03$_{\mathrm{+0.42}}$ & \textbf{12.06}$_{\mathrm{+0.45}}$ & 18.4 \\
    \makecell[c]{\textbf{Avg. $\Delta$}}        &  --   & -3.59 & \textbf{+2.27}           & 20.8 & --    & -4.86 & \textbf{+7.00}           & 25.5 & --    & -1.92 & \textbf{+2.19}           & 22.2 \\
    \bottomrule
  \end{tabular}}
\end{table*}

\subsection{Experimental Setup}
\label{sec:exp-setup}

\paragraph{Benchmark.}
To evaluate profile refinement in multi-domain, lifelong personalization settings, we construct a composite benchmark by integrating PersonaMem-v2~\citep{personamemv2} and MemoryCD~\citep{zhang2026memorycd}. We first derive task-agnostic user summaries from both datasets and extract semantic signatures capturing stable interests, preferences, aversions, and contextual constraints. We then match users one-to-one based on semantic compatibility, excluding pairs with explicit preference conflicts. For each matched pair, we construct a shared universal profile by interleaving and summarizing their PersonaMem and MemoryCD histories. Each composite user is evaluated with the same universal profile across 13 downstream tasks: nine PersonaMem-v2 conversational tasks and four MemoryCD tasks---item ranking, rating prediction, review-title generation, and review generation. For each task, we construct user-disjoint support, development, and evaluation sets, ensuring that users involved in policy induction do not appear in the held-out evaluation set. Detailed dataset statistics are provided in Appendix~\ref{app:benchmark-statistics}. We also evaluate the two original benchmarks separately and report the results in Section~\ref{app:separate-benchmark}.

\paragraph{Evaluation Metrics.}
We evaluate four-way choice questions with exact accuracy, item ranking with Hit@1, rating prediction with \emph{Rating-Score}, and both generation tasks with ROUGE-L. User profiles are constructed under the all-history memory setting, which summarizes user behavior from the complete historical context. Given the ground-truth rating $y$ and model prediction $\hat{y} \in \{1,2,3,4,5\}$, Rating-Score is defined as
$
\mathrm{S}(\hat{y}, y)=\max\left(0, 1 - \frac{|\hat{y}-y|}{4}\right).
$

Alongside task-level scores, we report context compression using the token ratio (TR), defined as the average refined-profile length divided by the average universal-profile length.

\paragraph{Implementation Details.}
In the main experiments, we use Gemini-2.5-Pro~\citep{gemini25} to generate a universal user preference from each user's raw history provided by the benchmark. Unless otherwise stated, both the meta learner and the preference refiner use Gemini-2.5-Pro. We set the support batch size to $b=20$ and the number of policy-update rounds to $T=5$, subject to our experimental budget. When more demonstrations are available, we select the support examples using the adaptive sampling method described in Appendix~\ref{sec:adaptive-sampling} to mitigate sampling bias. The universal-preference baseline directly passes the universal preference to the downstream model without refinement. The RAG baseline uses BM25~\citep{bm25}. For each query, we segment the universal profile at structural boundaries and further divide it into overlapping windows of 120 words with a 30-word overlap. We use the downstream task prompt as the retrieval query and rank all profile chunks with BM25 ($k_1=1.5$, $b=0.75$). The top eight chunks are restored to their original order and concatenated within an approximate 768-token profile budget ($TR\approx20\%$), which is comparable to the length of profiles refined by our method. The downstream models are {Qwen3-8B}~\citep{qwen3}, {DeepSeek-V4-Flash}~\citep{deepseekv4}, and {GPT-5-mini}~\citep{gpt5}.
We evaluate generalization with DeepSeek-V4-Flash as the meta model in Appendix~\ref{app:dpsk-meta-model}, provide further analyses beyond performance comparisons in Appendix~\ref{app:further-analysis}, and present case studies in Appendix~\ref{app:case_study}. Unless otherwise specified, Qwen3-8B is used as the default downstream model for these analyses.
 
\subsection{Main Results}

We evaluate whether \textsc{AlignXada} improves the performance--context trade-off across heterogeneous personalization tasks and downstream models. An effective preference refinement method should reduce preference context while preserving or improving downstream performance. We compare \textsc{AlignXada} with the raw universal-preference and RAG baselines across $39$ task--model cells.

The results in Table~\ref{tab:main-result} yield three observations.
\textbf{(1) \textsc{AlignXada} consistently improves the performance--efficiency trade-off across models and tasks.}
Across the $39$ task--model cells, \textsc{AlignXada} improves $33$, with an average gain of $+3.82$ points over the raw universal preference. All three downstream models improve on average: $+2.27$ points for Qwen3-8B, $+7.00$ for DeepSeek-V4-Flash, and $+2.19$ for GPT-5-mini. The gains also span task formats, covering all $12$ MemoryCD cells. Meanwhile, the refined profiles retain only $22.8\%$ of the original tokens, and performance gains are nearly uncorrelated with token ratio ($r=0.06$). Thus, \textsc{AlignXada} improves the utility of retained evidence rather than relying on longer profiles.
\textbf{(2) The advantage over RAG shows that task-oriented preference adaptation goes beyond query-level retrieval.}
\textsc{AlignXada} outperforms RAG in $36$ cells, ties in one, and underperforms it in only two, with an average margin of $+7.28$ points. In contrast, RAG reduces the raw-profile score by $3.46$ points on average. For example, on professional email with DeepSeek-V4-Flash, RAG lowers accuracy from $46.67\%$ to $35.83\%$, whereas \textsc{AlignXada} raises it to $59.17\%$. Retrieval may surface locally relevant records but fragment preferences whose relevance is indirect or distributed across interactions. \textsc{AlignXada} instead reorganizes the consolidated profile into a coherent decision context.
\textbf{(3) The optimal preference representation depends on both the task and the downstream model.}
For professional email, the changes are $-1.50$, $+12.50$, and $-0.67$ points for Qwen3-8B, DeepSeek-V4-Flash, and GPT-5-mini, respectively; for knowledge query, they are $+2.81$, $-0.06$, and $-1.58$ points. These sign reversals indicate that no single compression strategy is optimal for all downstream models, motivating the use of downstream feedback for policy induction. Despite this heterogeneity, all regressions remain within $1.58$ points, whereas the largest gain reaches $+14.77$ points.

\subsection{Results on Source-Native Benchmarks}
\label{app:separate-benchmark}

\begin{table}[t]
  \centering
  \caption{Task-level performance and token ratios on separate PersonaMem-v2 and MemoryCD (values $\times 100\%$).}
  \label{tab:separate-personamem}
  \adjustbox{width=0.95\linewidth}{
  \begin{tabular}{lcccc}
    \toprule
    \textbf{Task} & \textbf{Raw} & \textbf{RAG} & \textbf{\textsc{AlignXada}} & {TR$\downarrow$} \\
    \midrule
    \multicolumn{5}{c}{\textbf{Benchmark=PersonaMem-v2}} \\
    Chat Message        & 30.11 & 27.42 & \textbf{30.91} & 31.4 \\
    Creative Writing    & 31.13 & 27.13 & \textbf{32.78} & 37.0 \\
    Knowledge Query     & \textbf{49.92} & 43.98 & 48.71 & 43.3 \\
    Personal Email      & 30.93 & 29.83 & \textbf{32.39} & 40.4 \\
    Professional Email  & 31.09 & 27.72 & \textbf{33.83} & 38.7 \\
    Professional Writing& \textbf{28.46} & 24.73 & 27.66 & 51.6 \\
    Social Media Post   & 25.99 & 29.66 & \textbf{29.94} & 39.7 \\
    Translation         & 31.80 & 29.68 & \textbf{32.43} & 38.5 \\
    Trouble Consult     & 32.77 & 33.06 & \textbf{35.29} & 46.3 \\
    Avg. $\Delta$       & --    & -2.11 & \textbf{+1.30} & 40.8 \\
    \midrule
    \midrule
    \multicolumn{5}{c}{\textbf{Benchmark=MemoryCD}} \\
    Item Ranking        & 57.14 & 54.25 & \textbf{61.14} & 62.1 \\
    Rating Prediction    & 78.86 & 75.81 & \textbf{80.00} & 59.5 \\
    Review Title        & 13.47 & 14.32 & \textbf{14.36} & 44.7 \\
    Review Generation    & 13.72 & 14.04 & \textbf{14.46} & 25.8 \\
    Avg. $\Delta$       & -- & -1.19 & \textbf{+1.69} & 48.0 \\
    \bottomrule
  \end{tabular}}
\end{table}

The composite benchmark merges heterogeneous histories from PersonaMem-v2 and MemoryCD into a universal user profile, exposing each downstream task to both relevant and unrelated evidence. Although this setting reflects the noisy histories of lifelong agents, \textsc{AlignXada} may benefit mainly from removing cross-domain noise introduced by benchmark construction. We therefore evaluate \textsc{AlignXada} separately on the two source benchmarks, where profiles are built from more domain-coherent native histories, to test whether it remains effective from a cleaner starting point.

The results in Table~\ref{tab:separate-personamem} show two patterns.
\textbf{\textsc{AlignXada} remains effective on cleaner, source-native profiles.}
It improves the average primary metric by $1.30$ points on PersonaMem-v2, outperforming the raw profile on seven of nine tasks, and by $1.69$ points on MemoryCD, improving all four tasks. These gains indicate that \textsc{AlignXada} goes beyond removing cross-domain noise by using support-set feedback to induce preference representations better aligned with downstream requirements.
\textbf{Feedback-based adaptation provides a better performance--compression trade-off than query-level retrieval.}
\textsc{AlignXada} outperforms RAG on every task, whereas RAG reduces the average score by $2.11$ points on PersonaMem-v2 and $1.19$ points on MemoryCD. Meanwhile, \textsc{AlignXada} reduces the profiles to average token ratios of $40.8\%$ and $48.0\%$, respectively. This suggests that support-set feedback reorganizes user evidence into a compact representation that is more useful to downstream models than locally retrieved excerpts.

\begin{figure}[t]
    \centering
    \includegraphics[width=0.95\linewidth]{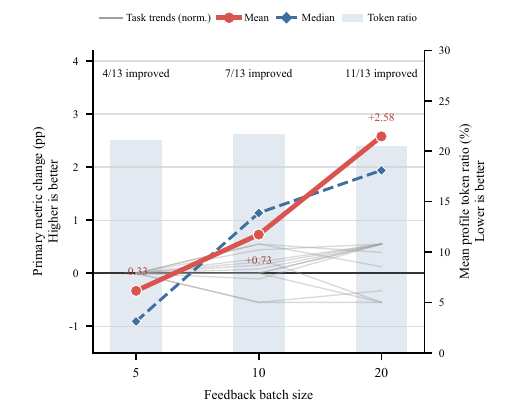}
    \caption{Effect of feedback batch size on the composite benchmark. Solid and dashed lines show the mean and median primary-metric changes, respectively. Gray lines show normalized task-level trends, bars show the mean profile token ratio, and annotations indicate the number of tasks outperforming the raw-profile baseline.}
    \label{fig:batch-size}
\end{figure}

\begin{figure}[t]
    \centering
    \includegraphics[width=0.95\linewidth]{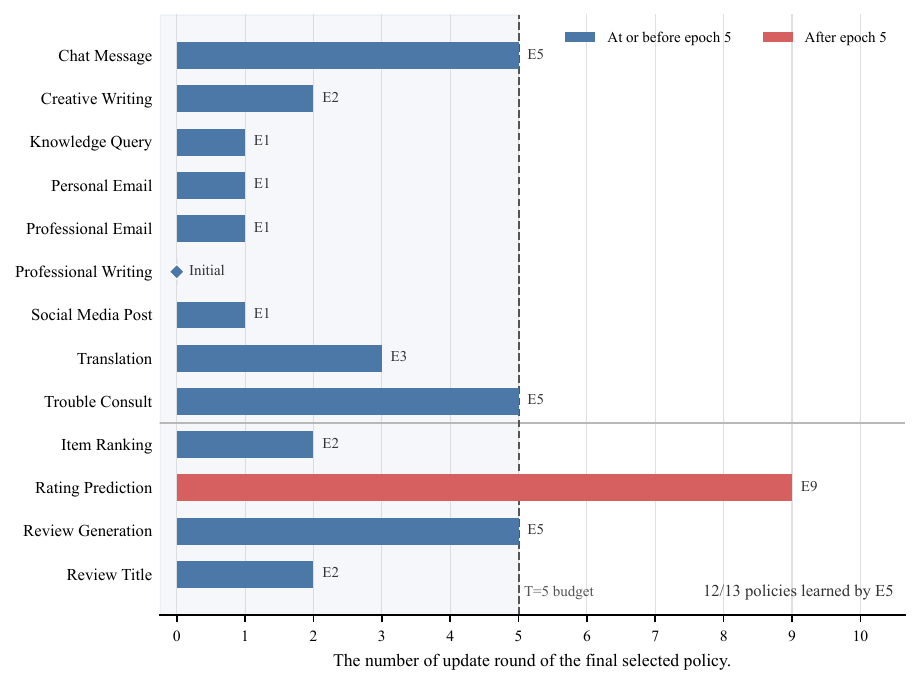}
    \caption{Update round of the final selected policy in the $T\in\{5,10\}$ diagnostic runs. Blue bars denote policies obtained by round five, red bars denote later policies. 
    }
    \label{fig:epoch}
\end{figure}

\subsection{Hyperparameter Analysis}

\paragraph{Impact of Support Batch Size $b$.}
Figure~\ref{fig:batch-size} compares feedback batch sizes $b\in\{5,10,20\}$ while keeping the induction pool and update-round budget fixed.
\textbf{(1) Larger feedback batches improve the reliability of policy induction.}
As $b$ increases from $5$ to $20$, the mean primary-metric change rises from $-0.33$ to $+2.58$ points, the median from $-0.91$ to $+1.94$ points, and the number of improved tasks from $4/13$ to $11/13$ ($7/13$ at $b=10$). Smaller batches make each update more sensitive to the sampled examples, whereas larger batches provide broader and more balanced feedback, leading to more stable policies.
\textbf{(2) The gains come from better feedback coverage rather than weaker compression.}
$b=20$ achieves the highest mean and median gains while producing the lowest mean token ratio ($20.5\%$, compared with $21.1\%$ at $b=5$ and $21.7\%$ at $b=10$). Thus, its advantage does not result from retaining more context. Instead, broader feedback helps distinguish recurring evidence-loss patterns from isolated failures and prioritize more useful preference evidence. We therefore use $b=20$ as the default, as it provides the best accuracy--compression trade-off among the evaluated settings. Larger batches are not evaluated because they exceed the context budget.

\paragraph{Impact of the Number of Update Rounds $T$.}
Figure~\ref{fig:epoch} shows when the final selected policy first appears in diagnostic runs with $T\in\{5,10\}$.
\textbf{(1) Effective refinement policies are typically learned within a few updates.}
Eight of the thirteen task-specific policies emerge by the second round, and twelve within five rounds, indicating that the meta-learner can quickly translate support-set feedback into effective policies.
\textbf{(2) Additional rounds offer task-dependent benefits.}
Rating prediction is the only task whose final policy first appears after round five, at round nine. Moreover, among the seven tasks for which $T=10$ performs better, six select policies generated within the first five rounds. Thus, most tasks converge early, while additional rounds mainly benefit challenging or unstable tasks. Balancing policy quality and inference cost, we use $T=5$ in the main experiments.

\subsection{Faithfulness Audit}
\label{sec:audit}

Performance and compression alone do not establish whether a refined preference can serve as a faithful substitute for the user universal preference. A refiner may reduce the context budget while removing the decisive preference evidence or introduce fictional traits. We therefore audit the refined preferences along two dimensions: \textbf{claim-level faithfulness} to the input profile and \textbf{preference evidence retention}. Experimental details are elaborated in Appendix~\ref{app:faithfulness}.

\begin{figure}[t]
    \centering
    \includegraphics[width=0.98\linewidth]{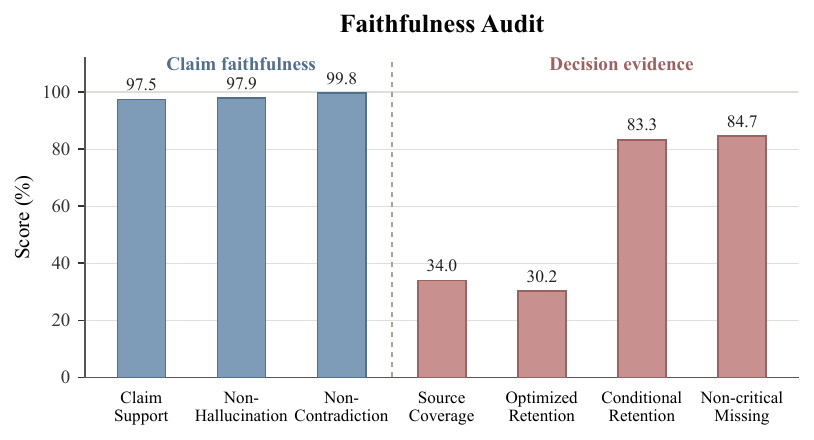}
    \caption{Faithfulness audit result on the PersonaMem-v2.}
    \label{fig:faithfulness-audit}
\end{figure}

We have two main observations about \textsc{AlignXada} from Figure~\ref{fig:faithfulness-audit}.
(1) \textbf{\textsc{AlignXada} is highly faithful to the source profile.}
Across all scenarios, $97.5\%$ of extracted claims are supported by the original profile, the non-hallucination rate is $97.9\%$, and the non-contradiction rate reaches $99.8\%$. These results show that the verbal policy does not turn the refiner into an unconstrained generator. Instead, \textsc{AlignXada} largely functions as a controlled compression-and-reorganization operator that preserves the factual content of the universal preference representation while adapting it for downstream use.
(2) \textbf{The main bottleneck is incomplete decision-evidence availability rather than profile hallucination.}
Only $34.0\%$ of benchmark preference evidence is recoverable from the original source profiles, while refined profiles retain $30.2\%$ overall. This small gap from the source-evidence ceiling suggests that many errors arise from insufficient evidence in the universal profile itself, rather than from the refiner deleting available evidence. Conditional on the original profile containing the golden preference, \textsc{AlignXada} preserves it in $83.3\%$ of cases, with a critical missing rate of $15.3\%$. Therefore, the audit provides a more precise interpretation of \textsc{AlignXada}'s downstream limitations: \textsc{AlignXada} can faithfully compress and adapt the profile it receives, but its performance ceiling depends on whether the universal profile contains the cross-topic evidence needed for the decision. The remaining headroom mainly lies in improving source-profile coverage and making decision-relevant evidence more consistently available to the refiner.

\section{Conclusion}
\label{sec:conclusion}

In this work, we study LLM personalization from the perspective of universal user preference interfaces. We propose \textsc{AlignXada}, which induces a reusable natural-language refinement policy through verbal reinforcement learning, improving the performance--budget trade-off while preserving source-supported claims and critical evidence. By enabling effective use of universal preferences, our work offers a new perspective on memory adaptation for lifelong personalized agents.

\section*{Limitations}

Despite its advantages, \textsc{AlignXada} has two main limitations. First, although our evaluation spans PersonaMem-v2 and MemoryCD across multiple task formats, both benchmarks are constructed or curated evaluation settings and may not fully capture the evolving, noisy interactions of real-world lifelong agents. Future work should develop broader datasets that provide both universal user preferences and diverse downstream task formats, enabling more comprehensive evaluation of task-specific preference adaptation. Second, \textsc{AlignXada} assumes that the support set used for policy induction is representative of the target task distribution. When the support set is biased or insufficiently diverse, the induced refinement policy may not generalize well. More effective sampling and support-set construction strategies remain important directions for future work.

\section*{Ethics Statement}

All experimental datasets used in this study are derived from previously published work and obtained either through official APIs or by synthetic construction based on these sources. All user-related information in the datasets has been anonymized, and no personally identifiable information is involved. We do not use any non-open-source data. All data are used solely for scientific research, rather than for commercial purposes or for profiling or decision-making about individuals, and their acquisition and use comply with relevant ethical guidelines and standards of academic integrity. All existing resources used in our experiments, including datasets, pretrained models, and APIs, are accessed and used in accordance with their original licenses and terms of use. The datasets and models we rely on were filtered and processed by their original authors prior to public release to mitigate potential ethical risks.

\bibliographystyle{plainnat}
\bibliography{custom}

\clearpage
\appendix

\section{Benchmark Statistics}
\label{app:benchmark-statistics}

\begin{table}[t]
\centering
\small
\caption{Statistics of the constructed composite benchmark.}
\label{tab:composite_statistics}
\adjustbox{width=\linewidth}{\begin{tabular}{lrrr}
\toprule
\textbf{Task} & \textbf{Support} & \textbf{Query} & \textbf{Total} \\
\midrule
Personal Email           & 40 & 84  & 210 \\
Professional Email       & 40 & 120 & 256 \\
Professional Writing     & 40 & 110 & 269 \\
Creative Writing         & 40 & 88  & 239 \\
Translation              & 40 & 111 & 264 \\
Trouble Consult          & 40 & 103 & 243 \\
Chat Message             & 40 & 103 & 226 \\
Social Media Post        & 40 & 88  & 217 \\
Knowledge Query          & 40 & 169 & 416 \\
Item Ranking             & 40  & 52  & 131 \\
Rating Prediction        & 40  & 52  & 131 \\
Review Title Generation  & 40  & 52  & 131 \\
Review Generation        & 40  & 52  & 131 \\
\midrule
\textbf{Total} & \textbf{520} & \textbf{1,184} & \textbf{2,864} \\
\bottomrule
\end{tabular}}
\end{table}

\begin{table}[t]
\centering
\small
\caption{Statistics of the PersonaMem-v2 benchmark.}
\label{tab:personamem_statistics}
\adjustbox{width=\linewidth}{\begin{tabular}{lrrr}
\toprule
\textbf{Task} & \textbf{Support} & \textbf{Query} & \textbf{Total} \\
\midrule
Personal Email          & 49 & 424 & 473 \\
Professional Email      & 50 & 485 & 535 \\
Professional Writing    & 63 & 453 & 516 \\
Creative Writing        & 52 & 457 & 509 \\
Translation             & 52 & 478 & 530 \\
Trouble Consult         & 49 & 452 & 501 \\
Chat Message            & 49 & 444 & 493 \\
Social Media Post       & 56 & 426 & 482 \\
Knowledge Query         & 61 & 661 & 722 \\
\midrule
\textbf{Total} & \textbf{481} & \textbf{4,280} & \textbf{4,761} \\
\bottomrule
\end{tabular}}
\end{table}

\begin{table}[t]
\centering
\small
\caption{Statistics of the MemoryCD benchmark.}
\label{tab:memorycd_statistics}
\adjustbox{width=\linewidth}{\begin{tabular}{lrrr}
\toprule
\textbf{Task} & \textbf{Support} & \textbf{Query} & \textbf{Total} \\
\midrule
Item Ranking & 160 & 400 & 560 \\
Rating Prediction & 160 & 400 & 560 \\
Review Title Generation & 160 & 400 & 560 \\
Review Generation & 160 & 400 & 560 \\
\midrule
\textbf{Total} & \textbf{640} & \textbf{1600} & \textbf{2240} \\
\bottomrule
\end{tabular}}
\end{table}

Tables~\ref{tab:composite_statistics}--\ref{tab:memorycd_statistics} summarize the task distributions used in our main and source-native evaluations. Since a user may contribute multiple targets and PersonaMem-v2 users do not necessarily have an example in every conversational scenario, the number of available records varies across tasks. In Table~\ref{tab:composite_statistics}, \emph{Support} denotes the fixed induction budget used for each task, \emph{Query} denotes the held-out instances used for final evaluation, and \emph{Total} denotes all constructed task records before runtime support and development subsampling; the latter therefore also includes candidate induction and development records not displayed in separate columns. Overall, the composite benchmark contains $2{,}864$ records across $13$ tasks. For the source-native evaluations, PersonaMem-v2 contributes $4{,}761$ examples across nine conversational tasks, while MemoryCD contributes $2{,}240$ examples across four recommendation and generation tasks. All experiments use user-disjoint induction and evaluation splits to prevent policy learning from exploiting evaluation-user histories.

\section{Adaptive Sampling for Policy Learning}
\label{sec:adaptive-sampling}

The meta learner has a limited context budget, so each policy update can use at most $b$ support examples even when a larger candidate pool is available. This makes the choice of support examples important. Fixed or randomly sampled batches may over-represent easy successes, persistent failures, or isolated regressions, leading to updates that are either too conservative or too reactive. To obtain a more balanced diagnostic view, we use adaptive diagnostic sampling, which selects support examples from multiple outcome-transition states.

In the main experiments, we reserve a candidate pool $\mathcal{C}^{(\tau)} =\{(P_{u_i},x_i,y_i)\}_{i=1}^{m}$ for adaptive support selection and a development set $D^{(\tau)}$ for final policy selection. Here, $m$ denotes the number of candidate examples available in the experimental split rather than a method-level budget. At each update, the sampler selects $b$ examples from $\mathcal{C}^{(\tau)}$ to form the support set $S^{(\tau)}$, where $b\leq m$. This allows each update to remain within the context budget while preserving broader coverage across users, topics, and difficulty levels. In deployment, the candidate-pool size is determined by the available task-specific interactions.

For each support example, we compare two binary outcomes: the outcome under the raw universal preference and the most recent outcome under the current refinement policy. This comparison assigns each example to one of four transition states:
\begin{itemize}
    \item \textsc{improved}: the raw profile is incorrect, but the refined profile is correct, indicating useful rewritten-profile representations;
    \item \textsc{regressed}: the raw profile is correct, but the refined profile is incorrect, exposing evidence that may have been removed or made less usable;
    \item \textsc{stable-success}: both the raw and refined profiles are correct, identifying safe compression behavior;
    \item \textsc{persistent-failure}: both the raw and refined profiles are incorrect, revealing evidence requirements that neither profile makes accessible to the downstream model.
\end{itemize}

At each update round, the sampler partitions the candidate pool by these transition states and allocates the support budget uniformly across them. If $b$ is not divisible by four, the remaining slots are assigned deterministically, with priority given to \textsc{improved} and \textsc{regressed} examples. If a transition state has too few examples, its unused slots are redistributed to the remaining states. This strategy provides the meta learner with both corrective signals from failures and stabilizing signals from successes in each update.

\begin{figure}
    \centering
    \includegraphics[width=\linewidth]{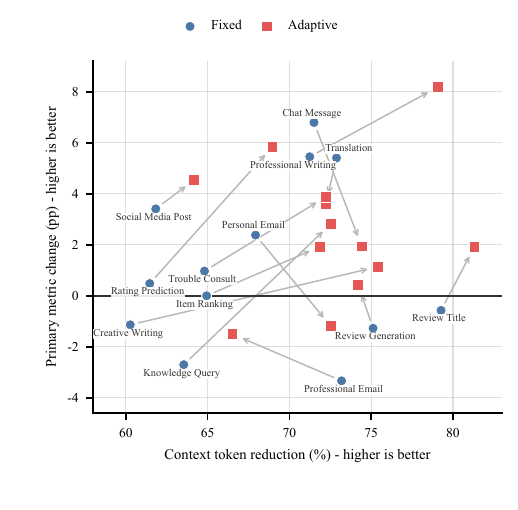}
    \caption{Fixed versus adaptive support sampling on the composite benchmark. Each arrow connects the same task under fixed sampling (circle) and adaptive sampling (square). The horizontal axis reports context-token reduction and the vertical axis reports the change in the task-specific primary metric; higher is better on both axes.}
    \label{fig:fixed}
\end{figure}

We compare fixed and adaptive sampling in Figure~\ref{fig:fixed} and make two observations.
\textbf{(1) Adaptive sampling improves the overall performance--compression trade-off rather than trading additional context for better predictions.}
The mean task-level gain in the primary metric increases from $+1.22$ points under fixed sampling to $+2.58$ points under adaptive sampling, while the number of tasks that improve over the raw-profile baseline rises from $7/13$ to $11/13$. At the same time, the mean context-token reduction increases from $68.3\%$ to $72.7\%$. Ten of the thirteen arrows move upward, ten move rightward, and eight move in both directions. Adaptive sampling therefore yields a broadly more favorable shift in performance--compression space by improving the selection of feedback examples used for policy induction, rather than by retaining longer refined profiles.
\textbf{(2) The primary benefit is the recovery of tasks for which fixed sampling induces unstable or incomplete policies.}
Adaptive sampling changes the gains for creative writing, knowledge query, review title generation, and review generation from negative to positive. It also increases the gain for rating prediction from $+0.48$ to $+5.85$ points, professional writing from $+5.45$ to $+8.18$ points, and trouble consultation from $+0.97$ to $+3.88$ points. These tasks benefit from repeatedly exposing the meta-learner to both regressions, which reveal decision-relevant evidence discarded by the current policy, and persistent failures, which reveal evidence that the policy still fails to surface. The effect is not uniform: fixed sampling remains stronger for chat message, personal email, and translation, while professional email remains below the raw-profile baseline. Adaptive sampling should therefore be interpreted as a more reliable strategy for allocating feedback across heterogeneous tasks, rather than as a guarantee of improvement on every task.

\section{Generalization Across Different Meta Models}
\label{app:dpsk-meta-model}

\begin{table}[t]
  \centering
  \caption{Performance and token ratio with DeepSeek-V4-Flash as the meta model, rewrite model, and the target model (values $\times 100\%$).}
  \label{tab:raw-rag-mepro}
  \adjustbox{width=\linewidth}{
  \begin{tabular}{lcccc}
    \toprule
    \textbf{Task} & \textbf{Raw} & \textbf{RAG} & \textbf{\textsc{AlignXada}-D} & {TR$\downarrow$} \\
    \midrule
    Chat Message        & 55.34 & 41.75 & 60.24 & 16.2 \\
    Creative Writing    & 59.09 & 54.55 & 68.71 & 23.1 \\
    Knowledge Query     & 82.90 & 71.01 & 79.16 & 30.7 \\
    Personal Email      & 45.24 & 39.29 & 48.81 & 16.9 \\
    Professional Email  & 46.67 & 35.83 & 56.67 & 24.6 \\
    Professional Writing& 50.00 & 40.91 & 52.13 & 26.5 \\
    Social Media Post   & 56.82 & 53.41 & 61.06 & 31.8 \\
    Translation         & 54.95 & 45.95 & 61.48 & 20.4 \\
    Trouble Consult     & 60.19 & 55.34 & 60.91 & 22.3 \\
    Item Ranking        & 69.23 & 78.85 & 82.42 & 33.7 \\
    Rating Prediction   & 74.04 & 76.44 & 77.83 & 26.2 \\
    Review Title        & 12.78 & 10.76 & 15.35 & 20.5 \\
    Review Generation   & 13.89 & 13.86 & 14.50 & 18.4 \\
    \bottomrule
  \end{tabular}}
\end{table}

To assess whether \textsc{AlignXada} generalizes beyond its default model configuration, we replace Gemini-2.5-Pro with DeepSeek-V4-Flash for policy induction and profile rewriting. We also use DeepSeek-V4-Flash as the downstream model, yielding an all-DeepSeek variant, \textsc{AlignXada}-D, while keeping the learning procedure and experimental protocol unchanged.

\textbf{\textsc{AlignXada} remains effective when all model roles use DeepSeek-V4-Flash, indicating that its gains are not specific to Gemini-2.5-Pro.}
\textsc{AlignXada}-D improves $12$ of the $13$ tasks over the raw universal preference, with an average gain of $+4.47$ points. It improves eight of the nine PersonaMem-v2 tasks, averaging $+4.22$ points, and all four MemoryCD tasks, averaging $+5.04$ points. The refined profiles retain only $23.95\%$ of the original tokens. These results demonstrate that \textsc{AlignXada} transfers across model families and task formats while maintaining a favorable performance--efficiency trade-off.

\textbf{Its consistent advantage over RAG suggests that this transfer stems from preference-level adaptation rather than model-specific retrieval behavior.}
\textsc{AlignXada}-D outperforms RAG on all $13$ tasks by $9.33$ points on average, whereas RAG falls $4.86$ points below the raw-profile baseline. Its largest gains over the raw profile occur on item ranking ($+13.19$), professional email ($+10.00$), and creative writing ($+9.62$), spanning ranking, classification, and open-ended generation. Thus, feedback-guided preference reorganization remains effective even when policy induction, rewriting, and downstream inference use the same model family.

\textbf{Different policy-induction and rewriting models nevertheless produce distinct performance--compression trade-offs.}
For the same DeepSeek-V4-Flash downstream model, the default Gemini-based configuration achieves an average gain of $+7.00$ points with a token ratio of $25.5\%$, compared with $+4.47$ points and $23.95\%$ for \textsc{AlignXada}-D. \textsc{AlignXada}-D therefore produces slightly more compact profiles but smaller performance gains, with knowledge query as its only regression ($-3.74$ points). These results distinguish framework generality from model interchangeability: \textsc{AlignXada} transfers across model families, but the choice of policy-induction and rewriting models still affects how effectively decision-relevant evidence is identified and retained.

\section{Diagnostic Analysis on PersonaMem-v2}
\label{app:further-analysis}

\subsection{Preference Source Robustness}
\label{app:exp-source}

\begin{table*}[t]
  \centering
  \caption{Ablation on the source of the universal preference (values $\times 100\%$).}
  \label{tab:source-ablation}
  \adjustbox{width=\linewidth}{
  \begin{tabular}{lcccccccccc}
    \toprule
    & \multicolumn{3}{c}{\textbf{Source A}} &
      \multicolumn{3}{c}{\textbf{Source B}} &
      \multicolumn{3}{c}{\textbf{Source C}} \\
    \cmidrule(lr){2-4}\cmidrule(lr){5-7}\cmidrule(lr){8-10}
    \textbf{Task} & Raw & \textsc{AlignXada} & TR($\downarrow$) & Raw & \textsc{AlignXada} & TR($\downarrow$) & Raw & \textsc{AlignXada} & TR($\downarrow$) \\
    \midrule
    Chat Message         & 28.37 & 28.64 & 36.8 & 30.19 & 29.73 & 16.4 & 30.11 & 30.91 & 31.4 \\
    Creative Writing     & 24.58 & 24.13 & 49.7 & 33.42 & 31.89 & 31.2 & 31.13 & 32.78 & 37.0 \\
    Knowledge Query      & 33.21 & 37.84 & 35.3 & 52.76 & 50.18 & 23.6 & 49.92 & 48.71 & 43.3 \\
    Personal Email       & 22.73 & 23.46 & 53.2 & 31.95 & 30.27 & 28.7 & 30.93 & 32.39 & 40.4 \\
    Professional Email   & 23.64 & 21.37 & 45.9 & 29.58 & 28.42 & 27.3 & 31.09 & 33.83 & 38.7 \\
    Professional Writing & 24.82 & 25.16 & 45.6 & 26.94 & 27.31 & 26.5 & 28.46 & 27.66 & 51.6 \\
    Social Media Post    & 26.35 & 26.78 & 45.2 & 30.67 & 29.14 & 23.8 & 25.99 & 29.94 & 39.7 \\
    Translation          & 25.91 & 27.32 & 41.5 & 34.28 & 34.76 & 21.4 & 31.80 & 32.43 & 38.5 \\
    Trouble Consult      & 21.46 & 21.83 & 34.7 & 26.37 & 26.92 & 30.1 & 32.77 & 35.29 & 46.3 \\
    Avg. $\Delta$        & --    & +0.61 & 43.1 & --    & -0.84 & 25.4 & --    & +1.30 & 40.8 \\
    \bottomrule
  \end{tabular}}
\end{table*}

To examine whether \textsc{AlignXada}'s behavior depends on the quality and format of the universal preference representation, we construct three sources of universal preference information on PersonaMem-v2. 
\textbf{Source A} summarizes a user's identity, background, interests, communication style, and other attributes using GPT-5~\citep{gpt5}, which is provided by the benchmark. We use A to test whether \textsc{AlignXada} can still improve a compact, curated persona preference that has already removed much of the conversational detail. 
\textbf{Source B} uses the persona's raw structured JSON record from the benchmark. This record is the generator-side persona artifact used to create the benchmark's preferences, conversation snippets, and answers. To avoid direct conversation leakage, we remove the benchmark conversation branch while retaining the structured persona and preference fields. We use B as a high-information structured-source stress test to measure whether a verbal refiner can replace hand-designed field selection when explicit preferences are available. 
\textbf{Source C} summarizes each user's raw chat history into a comprehensive natural language preference description using Gemini-2.5-Pro. Unlike A and B, C is therefore a history-derived preference rather than a persona-generation artifact. We use C as the default source in the main experiments because it best matches our target setting: a system first summarizes long user--assistant interactions into a rich but task-agnostic preference summary, and \textsc{AlignXada} then adapts that profile into a task-specific preference representation.

The results are shown in Table~\ref{tab:source-ablation}, from which we draw the following observations.
\textbf{(1) \textsc{AlignXada} works best as a task-specific refiner for natural-language preference profiles, rather than as a replacement for upstream profile construction.}
Both Source A and Source C improve after refinement, with absolute gains of $+0.61\%$ and $+1.30\%$, respectively, whereas Source B starts from the strongest raw baseline but decreases after refinement. This pattern suggests that \textsc{AlignXada} benefits from sources whose evidence is already expressed in natural language, while raw structured records may require a more field-aware transformation.
\textbf{(2) The gains on A and C come from reorganizing available preference evidence into a more task-oriented form.}
Source A has the lowest raw accuracy ($25.67\%$), because the dataset-provided expanded persona is compact and often lacks the full cross-topic evidence required by downstream tasks. \textsc{AlignXada} still improves it to $26.28\%$, showing that even a limited persona can benefit from task-specific reorganization. Source C starts from a stronger raw baseline ($32.47\%$), as the comprehensive profile contains richer user history, and \textsc{AlignXada} further improves it to $33.77\%$. This makes Source C the best match for our target setting: it is sufficiently evidence-rich for personalization while remaining in natural language, allowing the refiner to reliably compress and reorganize it.
\textbf{(3) The drop on B suggests a format mismatch rather than a lack of useful information.}
Source B's high raw accuracy ($32.91\%$) is expected, because it is the structured persona used to synthesize the benchmark and often contains explicit preference aligned with the query-required evidence. However, after refinement, accuracy drops to $32.07\%$. One likely reason is that many useful signals are stored as fine-grained structured fields or leaf values; rewriting them into a compact natural-language profile can remove exact decision evidence that the downstream model can directly read. From the compression perspective, this result also shows that strong compression alone is insufficient.
The best trade-off is achieved by Source C, where \textsc{AlignXada} obtains a positive gain while reducing the profile to $40.8\%$ of its original token length.

\subsection{Faithfulness Audit Details}
\label{app:faithfulness}

In Section~\ref{sec:audit}, we audit refined preferences along two dimensions: \textbf{claim-level faithfulness} to the input profile and \textbf{preference evidence retention}. The experimental setup and implementation details are as follows:

\paragraph{Claim-level faithfulness audit.}
For each unique raw--refined preference pair, we use DeepSeek-V4-Pro~\citep{deepseekv4} as the judge model to decompose the refined preference into atomic claims and label each claim with respect to the universal preference as \emph{supported}, \emph{contradicted}, \emph{not found}, or \emph{too vague}. Let $N_c$ denote the total number of atomic claims across all audited pairs. The three claim-level metrics---Claim Support, Non-Hallucination, and Non-Contradiction---are defined as follows:
\begin{equation}
\small
\begin{aligned}
\mbox{Claim Support}
&= \frac{\#\mathrm{supp.}}{N_c}, \\
\mbox{Non-Hallucination}
&= 1 - \frac{\#\mathrm{notf.} + \#\mathrm{cont.}}{N_c}, \\
\mbox{Non-Contradiction}
&= 1 - \frac{\#\mathrm{cont.}}{N_c}.
\end{aligned}
\end{equation}
Here, \emph{supported} indicates that a claim is explicitly stated or semantically entailed by the universal preference, \emph{not found} indicates unsupported new information, and \emph{contradicted} indicates inconsistency with the source. Claims labeled \emph{too vague} are included in $N_c$ but are not counted as supported. We define the judge prompt as follows:

\begin{tcolorbox}[colframe=gray!20!white, colback=gray!3!white, coltitle=black, fonttitle=\bfseries, title=Claim-Level Faithfulness Audit, breakable]
\small\ttfamily
Audit the refined profile against the original profile.\par\medskip
Labels:\par
- supported: the claim is explicitly stated or semantically entailed by the
original profile.\par
- contradicted: the claim conflicts with the original profile.\par
- not\_found: the claim adds specific information not present in the original
profile.\par
- too\_vague: the claim is so generic that it cannot be checked as a concrete
profile claim.\par\medskip
Return exactly one JSON object with this schema:\par
\{\par
\quad "claims": [\par
\quad\quad \{\par
\quad\quad\quad "claim": "atomic claim from the refined profile",\par
\quad\quad\quad "label": "supported | contradicted | not\_found | too\_vague",\par
\quad\quad\quad "evidence": "short supporting or contradicting evidence from the original profile, or empty string",\par
\quad\quad\quad "rationale": "brief reason"\par
\quad\quad \}\par
\quad ]\par
\}\par\medskip
Split the refined profile into concise atomic claims. Do not invent claims
that are not in the refined profile.\par\medskip
<OriginalProfile>\par
\{original\_profile\}\par
</OriginalProfile>\par\medskip
<RefinedProfile>\par
\{refined\_profile\}\par
</RefinedProfile>
\end{tcolorbox}

\paragraph{Preference-evidence audit.}
PersonaMem-v2 provides a ``preference'' field for each benchmark query, which we treat as ground-truth decision evidence. For each audited query, the judge determines whether this evidence is preserved in the universal preference and in the refined preference. 
Each preference is labeled as \emph{exact/specific}, \emph{semantically generalized}, \emph{missing}, or \emph{contradicted}, where the first two labels are counted as evidence \textbf{retained}. 
Let $N_e$ denote the number of audited query examples. We define four evidence-level metrics: Source Coverage, Refined Retention, Conditional Retention, and Non-Critical Missing: 
\begin{equation}
\small
\begin{aligned}
\text{Source Coverage}
&= \frac{\#\mathrm{source\ ret.}}{N_e}, \\
\text{Refined Retention}
&= \frac{\#\mathrm{refined\ ret.}}{N_e}, \\
\text{Conditional Retention}
&= \frac{\#\mathrm{source\ ret. \ \&\ refined\ ret.}}
        {\#\mathrm{source\ ret.}}, \\
\text{Non-Critical Missing}
&= 1 - \frac{\#\mathrm{critical\ missing}}
              {\#\mathrm{source\ ret.}}.
\end{aligned}
\end{equation}
Here, \emph{source ret.} indicates that the universal preference contains the ground-truth evidence, and \emph{refined ret.} indicates that the refined preference preserves it. A \emph{critical missing} case occurs when the universal preference retains the evidence but the refined preference is labeled \emph{missing}.  The judge prompt is defined as follows:

\begin{tcolorbox}[colframe=gray!20!white, colback=gray!3!white, coltitle=black, fonttitle=\bfseries, title=Preference-Evidence Audit, breakable]
\small\ttfamily
Audit whether each profile preserves the benchmark preference.\par\medskip
Labels:\par
- exact\_or\_specific: the profile explicitly preserves the preference or a
highly specific equivalent.\par
- semantic\_generalized: the profile preserves the preference at a broader but
still useful level.\par
- missing: the profile does not contain usable evidence for the preference.\par
- contradicted: the profile states the opposite or says the preference was
retracted/should not be used.\par\medskip
Return exactly one JSON object with this schema:\par
\{\par
\quad "original": \{\par
\quad\quad "label": "exact\_or\_specific | semantic\_generalized | missing | contradicted",\par
\quad\quad "evidence": "short evidence span from the original profile, or empty string",\par
\quad\quad "rationale": "brief reason"\par
\quad \},\par
\quad "optimized": \{\par
\quad\quad "label": "exact\_or\_specific | semantic\_generalized | missing | contradicted",\par
\quad\quad "evidence": "short evidence span from the refined profile, or empty string",\par
\quad\quad "rationale": "brief reason"\par
\quad \}\par
\}\par\medskip
<BenchmarkPreference>\par
\{benchmark\_preference\}\par
</BenchmarkPreference>\par\medskip
<OriginalProfile>\par
\{original\_profile\}\par
</OriginalProfile>\par\medskip
<RefinedProfile>\par
\{refined\_profile\}\par
</RefinedProfile>
\end{tcolorbox}

\section{Prompt Templates}

\subsection{Rewrite Prompt Template}
\label{app:rewrite-prompt}

The refiner receives the current policy $\phi_t$, the downstream task family description, and the original profile $P_u$.

\begin{tcolorbox}[colframe=gray!20!white, colback=gray!3!white, coltitle=black, fonttitle=\bfseries, title=Prompt Template for Preference Generation, breakable]
\small\ttfamily
Current rewrite pattern phi\_t:\par
\{current\_policy\}\par\medskip
Downstream task family:\par
\{task\_description\}\par\medskip
Original profile P\_u:\par
\{original\_profile\}\par\medskip
Rewrite the profile strictly according to phi\_t.\par\medskip
Requirements:\par
1. Execute phi\_t exactly, including its preservation, compression, abstraction, structure, and style rules.\par
2. Preserve factual faithfulness to P\_u; do not add unsupported facts, preferences, or assumptions.\par
3. Keep the rewrite query-agnostic and do not optimize it for a specific prompt, option, answer, or evaluation instance.\par
4. Output only the rewritten profile P*\_u.
\end{tcolorbox}

\subsection{Meta-Learner Prompt Template}
\label{app:meta-learner-prompt}

The meta learner receives the current policy and mini-batch feedback, then emits only the next policy.

\begin{tcolorbox}[colframe=gray!20!white, colback=gray!3!white, coltitle=black, fonttitle=\bfseries, title=Prompt Template for Policy Update, breakable]
\small\ttfamily
Meta prompt version:\par
\{prompt\_version\}\par\medskip
Downstream task family:\par
\{task\_description\}\par\medskip
Primary optimization metric:\par
\{primary\_metric\_name\}\par\medskip
Current pattern phi\_t:\par
\{current\_policy\}\par\medskip
Support-set error information (JSON):\par
\{feedback\_payload\}\par\medskip
Optimization objective:\par
- Improve future rewritten-profile performance for the same task family while updating only the rewrite policy, not the downstream answer behavior.\par
- Preserve or improve the primary metric first; reduce redundant context only after retaining decision evidence.\par
- Keep the policy query-agnostic so it can serve unseen examples rather than the current support prompts.\par
- Treat concrete user facts, constraints, sensitive or anti-stereotypical preferences, and privacy behavior as potential decision evidence unless the feedback shows they are harmful.\par\medskip
Use feedback:\par
- Prioritize regressions as evidence of omitted, distorted, or under-weighted signals; preserve behaviors that explain improvements and stable successes.\par
- Separate evidence loss from downstream confusion, ambiguity, or source-profile insufficiency before changing the policy.\par
- Balance latent evidence with ordinary concrete evidence such as roles, places, objects, language/register, named interests, and everyday preferences.\par\medskip
Constraints:\par
- Make the smallest useful update to phi\_t and avoid overfitting to one
example, user, topic, or surface form.\par
- next\_pattern must describe how to rewrite profiles, not how to answer the
downstream task.\par
- Return only the final JSON.
\end{tcolorbox}

\subsection{Structured Feedback Template}
\label{app:feedback-format}

The structured feedback record $E_t$ is serialized as a JSON list, where each entry corresponds to one support example in the mini-batch. 
By default, the feedback does not include the full raw preference; instead, the meta learner observes the rewritten profile, paired outcomes, scores, and task metadata.

\begin{tcolorbox}[colframe=gray!20!white, colback=gray!3!white, coltitle=black, fonttitle=\bfseries, title=Feedback Template, breakable]
\small\ttfamily
[\par
\quad\{\par
\quad\quad "idx": int,\par
\quad\quad "prompt": string,\par
\quad\quad "reference": string,\par
\quad\quad "optimized\_profile": string,\par
\quad\quad "original\_prediction": string,\par
\quad\quad "optimized\_prediction": string,\par
\quad\quad "original\_primary\_score": float,\par
\quad\quad "optimized\_primary\_score": float,\par
\quad\quad "score\_delta": float,\par
\quad\quad "original\_metrics": object,\par
\quad\quad "optimized\_metrics": object,\par
\quad\quad "meta\_summary": \{\par
\quad\quad\quad "transition": "improved | regressed | persistent\_failure | stable\_success",\par
\quad\quad\quad "scenario": string,\par
\quad\quad\quad "pref\_type": string,\par
\quad\quad\quad "topic\_query": string,\par
\quad\quad\quad "reference\_letter": string,\par
\quad\quad\quad "gold\_signal": string,\par
\quad\quad\quad "original\_prediction": string,\par
\quad\quad\quad "original\_signal": string,\par
\quad\quad\quad "optimized\_prediction": string,\par
\quad\quad\quad "predicted\_signal": string,\par
\quad\quad\quad "prediction\_changed": bool\par
\quad\quad\},\par
\quad\quad "original\_profile\_words": int,\par
\quad\quad "optimized\_profile\_words": int,\par
\quad\quad "word\_reduction": int\par
\quad\}\par
]
\end{tcolorbox}

\section{Policy Examples}

\subsection{Initial Policy}
\label{app:phi0}

All experiments start from the same task-agnostic initial policy:

\begin{tcolorbox}[colframe=gray!20!white, colback=gray!3!white, coltitle=black, fonttitle=\bfseries, title=Initial Policy, breakable]
\small\ttfamily
Rewrite the profile into concise, stable preference rules that preserve relevant user signals while reducing unnecessary detail.
\end{tcolorbox}

\subsection{An example of Induced Policies}
\label{app:policy}


\begin{tcolorbox}[colframe=gray!20!white, colback=gray!3!white, coltitle=black, fonttitle=\bfseries, title=Example Induced Policy, breakable]
\small\ttfamily
Goal: Produce a compact, faithful refined profile that lets a Qwen3-8B downstream answer a 4-way personal-email choice question grounded in the user's relationship context, tone, and recurring email phrasing, while preserving cross-scenario evidence that disambiguates the choice.\par\medskip
Preserve:\par
- Relationship roles and names of recurring email recipients.\par
- Tone, greetings, signoffs, and recurring phrasing by relationship.\par
- Lifestyle constraints and cross-scenario preferences that affect email decisions.\par\medskip
Compress:\par
- Narrative chit-chat from chat history.\par
- Duplicated preferences or long descriptions that do not change the email decision.\par\medskip
Avoid:\par
- Removing cross-scenario evidence or confusing the user's preferences with those of family members.\par
- Copying exact answer-text phrasings from chat history.\par\medskip
Output Style:\par
- Bulleted, with explicit field labels and relationship/scenario tags.\par
- No prose narrative.\par\medskip
Priority: When in doubt, preserve cross-scenario preference evidence over
local email-style fluff.
\end{tcolorbox}

\section{Case Study}
\label{app:case_study}

We select a representative user case, together with its raw universal preference and refined preference, as a case study. The raw universal preference leads the downstream model to an incorrect answer, whereas the \textsc{AlignXada}-refined profile leads it to the correct one. The user asks for calming activities or audio options after a long day. Among the four candidate responses, the gold answer recommends a tranquil Bach harpsichord suite or softly recorded Gregorian chant. With the raw universal preference, the downstream model selects a response centered on Japanese shakuhachi and koto music; after refinement, it selects the gold response. 
The boxes below show lightly formatted excerpts copied from the corresponding JSON outputs, with omitted material marked by ellipses.

\begin{tcolorbox}[colframe=gray!20!white, colback=gray!3!white, coltitle=black, fonttitle=\bfseries, title=Raw Universal Preference Excerpt, breakable]
\small
This user is a highly educated, articulate, and reflective individual, likely
an academic in the humanities, with a professional focus on history, theology,
and liturgy, particularly within the Anglican tradition. Their name is Elaine
H. Nakamura. They reside in or near Portland, Oregon, and are of East Asian,
specifically Japanese, descent. Elaine is deeply engaged in both her
professional life as a university lecturer and her personal life within a parish
community, where she is a lay leader and choir member.\par\medskip
\ldots\ The rhythms of the church calendar (Advent, feast days) are
significant markers in her life.\par\medskip
Her aesthetic sensibilities are refined and consistent. She appreciates:\par
- \textbf{Music:} She is a choir member, rehearsing and performing works by
composers like Tallis. She has a sophisticated appreciation for classical music
(Bach's \emph{Jesu, meine Freude}), choral traditions, and the emotional impact
of music in different settings. She also enjoys karaoke with friends.\par
- \textbf{Literature:} She is an avid reader, enjoying 19th-century novels like
\emph{Middlemarch}, mystery novels (P.D. James, Sarah Waters), and poetry.
Reading is a cherished ritual, often done in the evening with tea and a cozy
blanket.\par
- \textbf{Visual Arts \& Design:} She is drawn to art galleries, theater (both
scripted and improvisational), and the aesthetics of interior design
\ldots
\end{tcolorbox}

\begin{tcolorbox}[colframe=gray!20!white, colback=gray!3!white, coltitle=black, fonttitle=\bfseries, title=\textsc{AlignXada}-Refined Profile Excerpt, breakable]
\small
\textbf{Tone \& Style}\par
- \textbf{Formality:} Prefers a polite, formal, and articulate communication
style, even in personal emails.\par
- \textbf{Tone:} Warm, reflective, and slightly literary.\par
- \textbf{Vocabulary:} Uses precise, rich, sensory vocabulary (e.g., ``faint
patter of rain,'' ``ancient cadences'').\par
\ldots\par
\textbf{Content \& Interests}\par
- \textbf{Professional Identity:} University lecturer in Church History. This
is a central part of her identity and informs her worldview.\par
- \textbf{Core Academic Interests:} Liturgical practices and history (Anglican
tradition, medieval mystery plays, funeral liturgies); theology and its
intersection with culture; 19th-century parish library development.\par
- \textbf{Cultural \& Aesthetic Preferences:}\par
\quad - \textbf{Music:} Classical (Bach), choral traditions (Tallis).
Participates in a parish choir. Also enjoys karaoke with friends.\par
\quad - \textbf{Literature:} 19th-century novels (\emph{Middlemarch}), mystery
novels (P.D. James, Sarah Waters), poetry. Reading is a cherished ritual.\par
\quad - \textbf{Arts:} Enjoys art galleries, theater (scripted and improv), and
appreciates handmade crafts like greeting cards.
\end{tcolorbox}

This example shows that \textsc{AlignXada} can improve downstream behavior without introducing new user facts. The raw preference already contains the relevant evidence, but it presents decision-relevant musical and liturgical preferences alongside many other salient identity and cultural details. In this case, the raw model appears to overweight the user's Japanese heritage and selects the candidate mentioning shakuhachi and koto. The refined profile reorganizes the same source-supported information into task-relevant clusters: classical music, choral practice, Anglican parish life, 19th-century reading, and quiet evening rituals. This makes the Bach/Gregorian-chant candidate more directly supported than the Japanese-instrument distractor, while reducing the profile by $46.6\%$.

\end{document}